\pdfoutput=1

\documentclass[11pt]{article}
\usepackage{authblk}
\usepackage{acl}

\usepackage{times}
\usepackage{latexsym}
\usepackage{graphicx}
\usepackage{amsfonts}
\usepackage{booktabs}
\usepackage{tabularx}
\usepackage{natbib}
\usepackage[T1]{fontenc}

\usepackage[utf8]{inputenc}

\usepackage{microtype}
\usepackage{lipsum}
\usepackage{array}

\makeatletter
\renewcommand\outauthor{
    \begin{tabular}[t]{>{\centering}p{15cm}}
    \bf\@author
    \end{tabular}}
\makeatother

\newcommand\blfootnote[1]{%
  \begingroup
  \renewcommand\thefootnote{}\footnote{#1}%
  \addtocounter{footnote}{-1}%
  \endgroup
}
\title{Leveraging Emotion-specific Features to Improve Transformer Performance for Emotion Classification}

\author[1]{\textbf{Atharva Kshirsagar*}}
\author[1]{\textbf{Shaily Desai*}}
\author[1]{\textbf{Aditi Sidnerlikar}}
\author[1]{\textbf{Nikhil Khodake}}
\author[2]{\textbf{Manisha Marathe}}

\affil[1,2]{Department of Computer Engineering, PVG's COET, Affiliated to Savitribai Phule Pune University, India.}

\affil[1]{\texttt {\{atharvakshirsagar145, shaily.desai21,}}
\affil[1]{\texttt {{sidnerlikaraditi6, nikhilkhodake2002\} @gmail.com}}}
\affil[2]{\texttt {mvm\_comp@pvgcoet.ac.in }}

\date{} 

\begin{document}
\maketitle
\begin{abstract}
\blfootnote{*Equal Contribution}
This paper describes team PVG’s AI Club’s approach to the Emotion Classification shared task held at WASSA $2022$. This Track $2$ sub-task focuses on building models which can predict a multi-class emotion label based on essays from news articles where a person, group or another entity is affected. Baseline transformer models have been demonstrating good results on sequence classification tasks, and we aim to improve this performance with the help of ensembling techniques, and by leveraging two variations of emotion-specific representations. We observe better results than our baseline models and achieve an accuracy of $0.619$ and a macro F1 score of $0.520$ on the emotion classification task.
\end{abstract}

\section{Introduction}\label{into}
Rapid growth in the availability of human-annotated text documents has led to an increase in methodologies for tasks such as classification, clustering and knowledge extraction. A multitude of sources have enabled public access to structured and semi-structured data comprising of news stories, written repositories, blog content, among countless other roots of information. \citep{bostan-klinger-2018-analysis} showed that the task of emotion classification has emerged from being purely research oriented to being of vital importance in fields like dialog systems, intelligent agents, and analysis and diagnosis of mental disorders. 

Humans themselves sometimes find it tough to comprehend the various layers of subtlety in emotions, and hence there has been only a limited amount of prior research revolving around emotion classification. It has been noted that larger deep learning models can also find it quite challenging to fully grasp the nuances and underlying context of human emotion.

With the advent of Transformer \citep{NIPS2017_3f5ee243} models, there has been an increase in performance for emotion classification of text-based models. Most transformer-based language models \citep{devlin2018bert,raffel2019exploring,radford2018improving} are pretrained on various self-supervised objectives. Combining transformer based sentence representations with domain-specialised representations for improving performance on the specific task has been successfully used in across many NLP domains \citep{peinelt-etal-2020-tbert, poerner-etal-2020-e, zhang-etal-2021-combining}. Building on these foundations, we propose a similar approach to the task of Emotion classification.

In this paper, we posit a solution to the WASSA $2022$ Shared Task on Empathy Detection, Emotion Classification and Personality Detection, specifically Track-$2$, emotion classification. We propose a hybrid model where we combine information from various entities to create a rich final representation of each datapoint, and the observed results show promise in combining the Transformer output with the emotion-specific embeddings and NRC features. 

The rest of the paper is organized in the following manner: Section \ref{data} offers an overview into the dataset on Empathetic concern in news stories, Section \ref{method} goes in depth about our proposed methodology with subsections describing the individual constituent modules. Section \ref{exp} explains the experimental and training setup along with the baselines used; Section \ref{results} elucidates the observed results, and Section \ref{conclusion} concludes this study.

\section{Dataset} \label{data}
\begin{figure}[htp]
    \centering
    \includegraphics[scale=0.4]{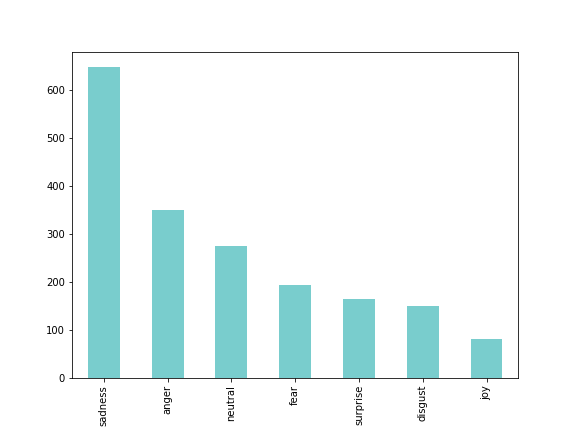}
    \caption{Distribution of the various classes among the Training Dataset}
    \label{fig:dataset}
\label{fig1}
\end{figure}

The dataset provided by the organizers consists of $1860$ essays in the training set, $270$ in the dev set and $525$ in the test set. Each of these essays has been annotated for empathy and empathy scores, distress and distress scores, emotion, personality feature and interpersonal reactivity features. Since this paper describes an approach only to the Emotion classification task, we shall only describe the data for said subtask. Each essay has been assigned an emotion class similar to classes in \citep{doi:10.1080/02699939208411068}. Table \ref{table1} provides a description on the training, validation and testing subset, and Figure \ref{fig1} shows the distribution of the training data among the various emotion classes.

\begin{table}[h!]
\centering
\begin{tabular}{cc}
\hline \textbf{Set} & \textbf{Essays}\\
\hline\hline
Training & 1860\\
Validation & 270\\
Testing & 525\\
\hline\hline
\textbf{Total} & 2655\\
\hline
\end{tabular}
\caption{\label{Data-Distribution}Total datapoints for every set}
\label{table1}
\end{table}

\section{Methodology}\label{method}

\subsection{RoBERTa}

We make use of the pretrained RoBERTa base model \citep{Liu2019RoBERTaAR} for this task. RoBERTa provides contextualized essay-level representations which can capture context sensitive information better than static representations. For each essay $\mathnormal{E}$ in our corpus, we obtain a $768$ dimensional representation $\mathnormal{R}$, encoded using the CLS token in the final hidden layer of the RoBERTa base model. We further process this representation $\mathnormal{R}$ with Linear and Dropout layers before concatenating it with our emotion-specific representations.
\begin{equation}
\mathnormal{R} = \mathnormal{RoBERTa}(\mathnormal{E}) \in \mathbb{R^{\mathnormal{d_1}}}
\end{equation}

\begin{figure}[t]
\centering
\frame{\includegraphics[width=0.45\textwidth,height=9cm]{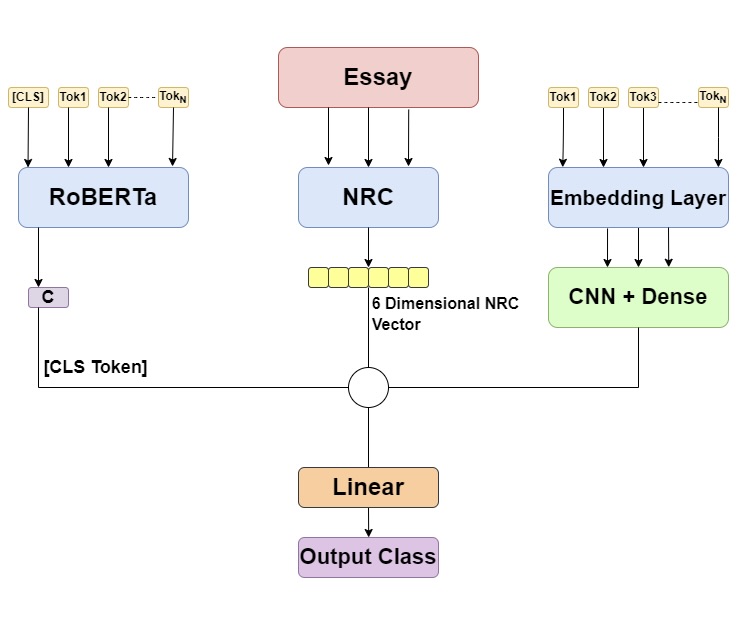}}
  \caption{Model Architecture}
\label{fig2}
\end{figure}

\subsection{Emotion-Enriched Word Embeddings(EWE)}

\citep{labutov-lipson-2013-embedding, bansal-etal-2014-tailoring},argue that the effectiveness of word embeddings is highly task dependent. To obtain word embeddings specific for emotion classification, we used the emotion-enriched embeddings from \citep{agrawal-etal-2018-learning}. The weight matrix was made by mapping the vocabulary from our dataset to the $300$ dimensional corresponding vector in the pre-trained embedding file. 
Each essay was mapped to the embedding matrix into a final representation shape of ($100$,$300$). This representation was passed through $2$ Conv1d and $2$ Maxpool layers to obtain a $16$ dimensional feature vector $\mathnormal{C} \in \mathbb{R^{\mathnormal{d_2}}}$.

\subsection{NRC Representation} \label{NRC}
The NRC emotion intensity lexicon \citep{mohammad-2018-word} is a collection of close to $\mathnormal{10,000}$ words associated with a distinct real valued intensity score assigned for eight basic emotions. Incorporating this lexicon in classification tasks has been proven to boost performance \citep{kulkarni-etal-2021-pvg}. Of the $\mathnormal{8}$ basic emotions in the lexicon, $\mathnormal{6}$ emotions-anger, joy, sadness, disgust, fear and surprise coincide with the given dataset and hence lexical features for only these features were considered. 
For every essay in the dataset, we calculate the value for one emotion by summing the individual scores for

\begin{table*}[!htb] 
\begin{tabular*}{\textwidth}{c @{\extracolsep{\fill}} ccccc}
\toprule
\textbf{Model} &  \multicolumn{2}{c}{\textbf{Accuracy}} & \multicolumn{2}{c}{\textbf{ Macro-F1 Score}}\\
\midrule
{}   & Training   & Validation    & Training   & Validation\\
Vanilla RoBERTa   &  0.601 & 0.540   & 0.513  & 0.452\\
RoBERTa + EWE   &  0.684 & 0.608   &  0.561 & 0.499\\
\midrule
\textbf{RoBERTA + NRC + EWE} & \textbf{0.693} & \textbf{0.619} & \textbf{0.618} & \textbf{0.520}\\
\bottomrule
\end{tabular*}
\caption{\label{performance-comparison}Resulting metrics on baseline models as compared to our methodology}
\label{table2}
\end{table*}

every word $\mathnormal{W}$ in the essay that occurs in the NRC lexicon. We then create a six dimensional vector $\mathnormal{N}$ corresponding to that essay which consists of the scores of the emotions in our dataset.\\

For a datapoint $\mathnormal{E}$, the six values of $\mathnormal{S_{emotion}}$ and the feature vector $\mathnormal{N}$ was constructed in the following manner:\\

\begin{equation}
\mathnormal{S_\mathnormal{emotion}} = \sum\mathnormal{W_\mathnormal{emotion}}   (\mathnormal{W} \in \mathnormal{E})
\end{equation}

\begin{equation}
N = [S_\mathnormal{anger};S_\mathnormal{joy};.....;S_\mathnormal{surprise}] \in \mathbb{R^{\mathnormal{d_3}}}
\end{equation}

\subsection{Combined Representation and Classification}
The feature vectors obtained from the RoBERTa ($\mathnormal{R}$), Emotion-Enriched Embeddings ($\mathnormal{C}$) and NRC ($\mathnormal{N}$)were concatenated to obtain the final representation ($\mathnormal{F}$).

\begin{equation}
\mathnormal{F} = [\mathnormal{R} ; \mathnormal{C} ; \mathnormal{N}] \in \mathbb{R^{\mathnormal{d_1 + d_2 + d_3}}}
\end{equation}

This representation is then passed through a single Linear layer with the Softmax activation. Figure \ref{fig2} depicts the model architecture in detail.

\section{Experimental Setup} \label{exp}
\subsection{Data Preparation}
Standard text cleaning steps like removing numbers, special characters, punctuation, accidental spaces, etc. were applied to each essay in the corpus. Stopwords were removed using the {\fontfamily{qcr}\selectfont nltk
} \citep{10.3115/1118108.1118117} library. Every essay was tokenized to a maximum length of $100$, and essays larger than this length were truncated. No standardization was done in the case of NRC scores, as we wanted to feed our model a vector of raw emotion-intensity scores for each of the six emotions considered in our NRC representation.

\subsection{Training Setup}
We used the pretrained 'roberta-base' model from the Huggingface {\fontfamily{qcr}\selectfont Transformers
}\footnote{\url{https://huggingface.co/transformers/}} library. All other modules used in our methodology were built using PyTorch. As observed by \citep{kulkarni-etal-2021-pvg}, we also found that the Hyperbolic Tangent(Tanh) activation function worked better than ReLU, and hence we used the Tanh activation for all layers in our model.  The model was trained using an AdamW optimizer \citep{loshchilov2018decoupled} with a learning rate of $0.001$ and beta values set to $\beta_1 = 0.9$, $\beta_2 = 0.99$ and the loss used was cross entropy loss. Additionally, early stopping was used if the validation loss does not decrease after 10 successive epochs. The batch size was set to $64$ for both Baseline models as well as the proposed model. A single Nvidia P100-16GB GPU provided by Google Colab was used to train all models.

\subsection{Baselines}
Our goal in this work is to examine if concatenating emotional-specific features to pre-existing transformer models leads to an increase in the emotion classification performance of these models. Hence, we compare our proposed methodology to the vanilla RoBERTa model, as well as RoBERTa + Ewe for the emotion classification subtask.

\section{Results and Discussion} \label{results}
The results for the emotion prediction task on the validation set are given in Table \ref{table2}. There was no use of validation data during the training process, and the provided validation data was used as unseen testing data to benchmark the models.The official metric for Track $2$ of the shared task was the macro F1 score. To ensure fair comparison, the validation set results have been averaged over $3$ runs for each model. 
The proposed model shows a $7$\% increase in macro F1 scores and $8$\% increase in accuracy over the vanilla RoBERTa model. The proposed model also shows the effectiveness of adding the NRC representations described in section \ref{NRC} as it performs slightly better than the RoBERTa + Emotion Enriched word embeddings model. We attribute this increase in performance to the task-specific representations of essays used in our system. 
During the training process, it was observed that the performance of all models was highly susceptible to how they were initialized, and we received a large range of results across different seeds. As a result, a true assessment of our method can only be made in comparison to baseline models with the same seed, as we have done in this study.

\section{Conclusion} \label{conclusion}
The goal of this study was to examine and enhance the performance of transformer models using only the Empathetic Concern in News Stories dataset that was provided to us, with the prospective of testing our method on a bigger dataset in the future. We proposed a model ensemble which combined the transformer feature vector with the emotion-intensive word embeddings along with the word-specific features obtained from the NRC lexicon. We demonstrate results that outperform the baseline vanilla RoBERTa model, and attest that combining domain-specific features can indeed improve performance on a task as involute as emotion classification. 

\bibliography{custom}
\bibliographystyle{acl_natbib}
\end{document}